\pdfoutput=1

\documentclass[11pt]{article}

\usepackage[final]{acl}

\usepackage{times}
\usepackage{latexsym}
\usepackage{enumitem}
\usepackage{amsmath}

\usepackage[T1]{fontenc}

\usepackage[utf8]{inputenc}

\usepackage{microtype}

\usepackage{inconsolata}

\newcommand{\ie}{\emph{i.e., }}
\newcommand{\eg}{\emph{e.g., }}

\usepackage{multirow}
\usepackage{graphicx}
\usepackage{booktabs}
\usepackage{float}

\title{Describe-then-Reason: Improving Multimodal Mathematical \\ Reasoning through Visual Comprehension Training}

\author{Author 1 \and ... \and Author n \\
        Address line \\ ... \\ Address line}
\author{Mengzhao Jia$^1$, {\bf Zhihan Zhang}$^1$, {\bf Wenhao Yu}$^2$, {\bf Fangkai Jiao}$^3$, {\bf Meng Jiang}$^1$ \\   $^1$University of Notre Dame $^2$Tencent AI Seattle Lab  $^3$Nanyang Technological University\\  $^1$\texttt{\{mjia2, zzhang23, mjiang2\}@nd.edu} \\ $^2$\texttt{wenhaowyu@global.tencent.com} $^3$\texttt{fangkai002@e.ntu.edu.sg}  }

\begin{document}
\maketitle
\begin{abstract}
Open-source multimodal large language models (MLLMs) excel in various tasks involving textual and visual inputs but still struggle with complex multimodal mathematical reasoning, lagging behind proprietary models like GPT-4V(ision) and Gemini-Pro. 
Although fine-tuning with intermediate steps (\ie rationales) elicits some mathematical reasoning skills, the resulting models still fall short in visual comprehension due to inadequate visual-centric supervision, which leads to inaccurate interpretation of math figures.
To address this issue, we propose a two-step training pipeline VCAR, which emphasizes the \textbf{V}isual \textbf{C}omprehension training in \textbf{A}ddition to mathematical \textbf{R}easoning learning.
It first improves the visual comprehension ability of MLLMs through the visual description generation task, followed by another training step on generating rationales with the assistance of descriptions. 
Experimental results on two popular benchmarks demonstrate that VCAR substantially outperforms baseline methods solely relying on rationale supervision, especially on problems with high visual demands.
\end{abstract}

\section{Introduction}
Open-source multimodal large language models (MLLMs), exemplified by models such as LLaVA~\cite{DBLP:journals/corr/abs-2310-03744} and Mini-GPT4~\cite{DBLP:journals/corr/abs-2304-10592}, has showcased impressive reasoning capabilities across tasks involving both textual and visual inputs such as visual question answering~\cite{DBLP:conf/cvpr/NamHK17,DBLP:journals/corr/abs-2310-10942} and multimodal dialogue~\cite{DBLP:journals/corr/abs-2308-03349}.
Despite these advancements, when addressing a more complex task of multimodal mathematical reasoning~\cite{DBLP:journals/corr/abs-2310-02255}, these open-source models are far lagged behind proprietary MLLMs like GPT-4V(ision)~\cite{gpt4v} and Gemini-Pro~\cite{DBLP:journals/corr/abs-2403-05530}. Such performance gap in solving math-related multimodal problems hinders their application potential in fields such as educational content generation~\cite{kasneci2023chatgpt} and statistical data analysis~\cite{DBLP:journals/corr/abs-2306-06031}.
Notably, specific deficiencies have been identified in these open-source models, including suboptimal performance in step-wise reasoning processes~\cite{DBLP:journals/corr/abs-2312-01598, DBLP:journals/corr/abs-2311-17076} and a tendency to overly rely on textual inputs while falling short in processing visual information~\cite{zhang2024mathverse}.

\begin{figure}[!t]
    \centering
    \includegraphics[width=\linewidth]{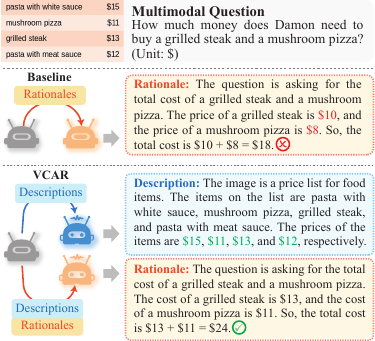}
    \vspace{-0.6cm}
    \caption{VCAR trains the model with an additional visual comprehension task in addition to the mathematical reasoning training, avoiding errors caused by inaccurate visual understanding.
    In contrast, the baseline method solely trained on rationales fails to correctly answer the question. It demonstrates the need for specified visual comprehension training.}
    \vspace{-0.6cm}
    \label{fig:intro}
\end{figure}

Noticing this gap, great research interests have been dedicated to enhancing the multimodal reasoning ability of open-source MLLMs. A popular direction is step-wise supervised fine-tuning, which involves training with supervision signals of intermediate reasoning steps (\textit{Rationales}) collected from language-based teacher models (\ie proprietary models such as GPT-4)~\cite{DBLP:journals/corr/abs-2305-03453,DBLP:conf/nips/ZhengYTZY23, DBLP:journals/corr/abs-2312-03052}. 
Although the collected rationales help elicit some reasoning ability in the MLLM, they contain limited content related to the visual perception of math figures. Such inadequate supervision limits the model's ability to comprehend math figures effectively, and eventually leads to the performance being restricted by inaccurate visual comprehension such as the example in Figure~\ref{fig:intro}.
While certain visual information could be acquired through tools such as visual question answering~\cite{DBLP:conf/nips/ZhengYTZY23} and object detection~\cite{DBLP:journals/corr/abs-2312-03052}, it is merely injected as supplementary to assist in composing coherent rationales, rather than being leveraged independently to teach models on how to recognize and understanding the visual content. 



To address the aforementioned issue, we propose to improve the multimodal mathematical reasoning ability of MLLMs by emphasizing the importance of visual comprehension training.
We introduce a novel two-step training pipeline that highlights \textbf{V}sual \textbf{C}omprehension training in \textbf{A}ddition to mathematical \textbf{R}easoning learning, dubbed as VCAR. 
Concretely, similar to the pre-training of MLLMs~\cite{DBLP:conf/nips/LiuLWL23a} that utilizes image description for initial alignment, we employ an image description generation task to equip the MLLM with enhanced visual understanding capability, thereby generating high-quality descriptions to assist the subsequent development of mathematical reasoning abilities.
Next, we train the MLLM to produce rationales based on the descriptions. With the text-form context provided by image descriptions, we isolate the training on mathematical reasoning skills from the demand of visual understanding.
This two-step approach systematically facilitates both the visual comprehension and mathematical reasoning faculties of the MLLMs, ensuring a balanced development of multimodal mathematical reasoning capabilities.

To acquire supervision signals for the aforementioned two-step training, we utilize Gemini-Pro to collect: (1) descriptive contents, used for comprehending the image, and (2) rationales, focusing on reasoning the answer.
Given the distinct objectives of each training step, we propose to optimize separate parameters tailored to each aspect to avoid imbalanced or disturbed learning. We adopt two Low-Rank Adaptation (LoRA)~\cite{DBLP:conf/iclr/HuSWALWWC22} modules to separately enhance visual understanding and mathematical reasoning abilities without the need for re-optimizing all model parameters.


To verify the effectiveness of our proposed VCAR training, we conduct experiments on two popular benchmarks MathVista~\cite{DBLP:journals/corr/abs-2310-02255} and MathVerse~\cite{zhang2024mathverse}, where VCAR outperforms all baseline methods without explicit visual understanding training. Specifically, VCAR achieves significant improvements on math problems that demand high visual understanding ability, \textit{i.e.}, $34.3\%$ and $13.3\%$ relative improvements on ``visual-only'' and ``visual-dominant'' categories on MathVerse.
Such improvements are consistent across two base MLLMs, which demonstrates the universality of VCAR.
Further ablation studies show that training in both visual comprehension and mathematical reasoning is crucial for performance improvement, and the two-step training pipeline is also indispensable in maximizing training efficacy.
Our main contributions are as follows:

\begin{itemize}
[noitemsep,topsep=1pt,parsep=1pt,partopsep=0pt]
    \item We highlight the significance of supervising the visual comprehension capability in improving multimodal mathematical reasoning, yet often neglected by existing methods that concentrate on the training of reasoning capacity.
    \item We introduce a two-step training pipeline named VCAR, which separates the training for visual comprehension and mathematical reasoning skills. 
    \item We conduct extensive experiments on two benchmarks, verifying the superiority of VCAR over various baselines. We also investigate the factors that contribute to the improvement. We released the codes and the collected supervision signals to facilitate other researchers in this community\footnote{\url{https://anonymous.4open.science/r/Describe-then-Reason-0650}.}.
\end{itemize}

\begin{figure}[!t]
    \centering
    \includegraphics[width=\linewidth]{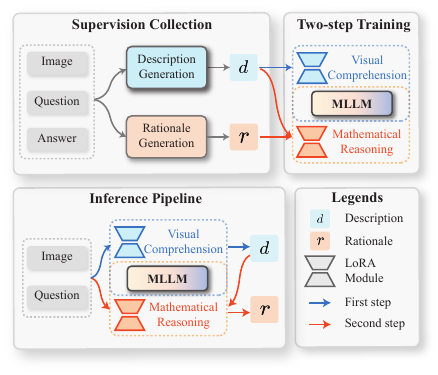}
    \caption{Illustration of the proposed method. VCAR consists of two components, namely supervision collection and two-step training. The inference pipeline of VCAR is also demonstrated.}
    \label{fig:case}
\end{figure}

\section{Related Work}

\subsection{Multimodal Large Language Models}
In light of the recent flourishing of large language models (LLMs), integrating multimodal capabilities into LLMs to derive MLLMs has attracted intense attention~\cite{InternLM-XComposer,Flamingo}. The architecture of contemporary MLLMs typically consists of a language foundation model (\eg LLaMA~\cite{LLaMA} or Vicuna~\cite{Vicuna}), a visual encoder (\eg Vision Transformer~\cite{ViT}), and a cross-modal alignment module. 
The development of MLLMs initiates with a cross-modal alignment training stage~\cite{DBLP:conf/nips/LiuLWL23a}, which utilizes paired image-description data to align the encoded visual features with the linguistic representation space of the LLM. This is followed by instruction tuning, where the model learns to interpret and execute instructions~\cite{DBLP:journals/corr/abs-2310-03744}. 
A prevalent method involves freezing the visual encoder's parameters during both training stages~\cite{MM1,DBLP:journals/corr/abs-2402-07865}. This practice, however, restricts the MLLMs' ability to fully grasp visual information~\cite{guan2023hallusionbench}. To mitigate this, our paper focuses on explicitly visual-centric supervision fine-tuning to specialize the training for visual comprehension capabilities.

\subsection{Multimodal Reasoning}
The boom of MLLMs has aroused researchers’ interest in multimodal reasoning tasks~\cite{you-etal-2023-idealgpt, Plug-and-Play, ViperGPT}. One popular direction is fine-tuning MLLMs with the supervision of intermediate reasoning steps (Rationales).
Existing work in this area utilizes rationales either authored by humans~\cite{DBLP:journals/corr/abs-2302-00923} or collected from proprietary large language models (LLMs)~\cite{DBLP:conf/emnlp/LinLMC23,DBLP:journals/corr/abs-2306-14122,DBLP:conf/emnlp/LinLMC23},  enhancing the model's reasoning capacity beyond direct answer learning.
Notably, DDCoT~\cite{DBLP:conf/nips/ZhengYTZY23} employs LLM to decompose the original question into sub-questions. Similarly, T-Sciq~\cite{DBLP:journals/corr/abs-2305-03453} and VPD~\cite{DBLP:journals/corr/abs-2312-03052} generate a question-resolving plan in natural language and programming code formats, respectively, thereby facilitate problem-solving.
Despite these remarkable achievements, they primarily concentrate on supervising the training from a language-based reasoning aspect. Little effort is devoted to enhancing visual comprehension. This lack of visual comprehension supervision can lead to sub-optimal performance in overall multimodal reasoning capabilities.

\section{Method}
To tackle the issue of inadequate visual comprehension training in multimodal mathematical reasoning, we propose VCAR, a two-step training pipeline designed to enhance both visual comprehension and mathematical reasoning abilities. 
In the first step, MLLMs undergo training in the image description generation task to bolster visual comprehension. Subsequently, the second training step focuses on mathematical reasoning improvement through description-assisted rationale generation.
In this section, we first introduce the process of collecting supervision signals for each learning objective, followed by the detailed two-step training pipeline.


\subsection{Problem Formulation}
Suppose that we have a set of $N$ multimodal training samples $\mathcal{I}= \{x_i\}_{i=1}^N$. Each multimodal sample $x_i \in \mathcal{I}$ consists of an image $v_i$, a textual question $q_i$, and a gold answer $y_i$.  We aim to train the model to learn a mapping function $\mathcal{F}: (v_i, q_i) \rightarrow y_i$.
The index (\ie $i$) of each training sample is omitted for simplicity in the following sections.

\subsection{Supervision Collection}
\label{sec:supervision_collection}
The high cost of human annotation leads to an absence of fine-grained supervision signals such as rationales, which poses challenges to the training of reasoning ability.
Noticing the advanced reasoning and sophisticated language generation capacity of proprietary LLMs or MLLMs, a prevalent practice is to collect the desired signals from them~\cite{DBLP:conf/acl/HsiehLYNFRKLP23, DBLP:conf/acl/ShridharSS23}. 
Inspired by this, we gather supervision signals to carry out training from these proprietary MLLMs. Concretely, we employ the Gemini-Pro and GPT-4V(vision) models and design distinct prompts to acquire two types of supervision signals.


%
\noindent \textbf{Rationale Generation}.
\label{sec:supervision}
Utilizing rationales to assist the learning of mathematical reasoning has shown advantages over mere answer supervision in a series of studies~\cite{DBLP:journals/corr/abs-2312-09241,MAmmoTH}. Motivated by this, we incorporate rationales into the training of MLLM's mathematical reasoning capability. 
A popular approach is to leverage prompts like ``\textit{Let's think step by step}'' to elicit the generation of intermediate reasoning steps. However, the proprietary models sometimes fail to correctly answer certain questions, leading to rationales with erroneous and inaccurate expressions, thereby damaging the training~\cite{mathgenie}. To alleviate this, we provide the ground truth answer when generating the rationale to enhance accuracy, similar to~\citet{DBLP:journals/corr/abs-2210-06726}. Concretely, we construct the rationale generation prompt for each sample as $p_r = (I_r,v,q,y)$, where $I_r$ represents the rationale generation instruction. The detailed $p_r$ is delineated below. By feeding $p_r$ into the proprietary models, we get a rationale $r$ for each sample. 

\begin{figure}[!h]
    \centering
    \includegraphics[width=\linewidth]{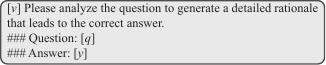}
    \vspace{-0.5cm}
    \label{fig:case}
\end{figure}



\noindent \textbf{Description Generation}.
To impose explicit supervision on the MLLM's visual comprehension ability, we additionally construct a visual description $d$ for each image $v$, besides the collection of the rationale $r$.
One intuitive way to implement this is to directly prompt proprietary MLLMs to describe the given image, which, however, presents two primary drawbacks.
Firstly, simply prompting MLLMs to exhaustively enumerate all details in the image often yields lengthy descriptions, thereby introducing unnecessary complexity for the description generation training and the following mathematical reasoning process. 
Besides, given the imperfect visual perception of MLLMs~\citep{guan2023hallusionbench,jing2023faithscore}, generating the description without any references is likely to introduce erroneous information, which would also affect the subsequent learning process adversely.

To ameliorate these weaknesses, we include the corresponding question and the correct answer as guidance when acquiring the visual description, which not only emphasizes the description's relevance to the question but also avoids potential errors.
In formulation, the visual description $d$ is obtained using the prompt $p_d = (I_d,v,q,y)$, in which $I_d$ is an instruction asking the model to describe the image regarding the current question and answer. Detailed prompt $p_d$ is shown as follows:
\begin{figure}[h]
    \centering
    \includegraphics[width=\linewidth]{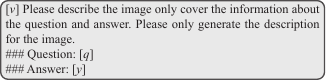}
    \label{fig:case}
\end{figure}
\begin{table*}[!t]
  \centering
  \resizebox{\textwidth}{!}{
    \begin{tabular}{l|c|ccccc|c|ccccc}
    \toprule
    \multirow{2}[4]{*}{\textbf{Method}} & \multicolumn{6}{c|}{\textbf{MathVista}}       & \multicolumn{6}{c}{\textbf{MathVerse}} \\
\cmidrule{2-13}          & \textbf{ALL} & \textbf{FQA} & \textbf{GPS} & \textbf{MWP} & \textbf{TQA} & \textbf{VQA} & \textbf{ALL} & \textbf{TD} & \textbf{TL} & \textbf{VI} & \textbf{VD} & \textbf{VO} \\
    \midrule
    \multicolumn{13}{c}{\textit{Proprietary MLLMs (Zero-shot)}} \\
    \midrule
    Gemini-Pro~\cite{DBLP:journals/corr/abs-2403-05530} & 47.7  & \textbf{48.3 } & 35.1  & 50.5  & \textbf{65.8 } & \textbf{42.5 } & 22.3  & 27.6  & 23.7  & 19.4  & 20.3  & 20.5  \\
    GPT-4V~\cite{gpt4v} & \textbf{49.9 } & 43.1  & \textbf{50.5 } & \textbf{57.5 } & 65.2  & 38.0  & \textbf{38.3 } & \textbf{52.1 } & \textbf{40.9 } & \textbf{34.9 } & \textbf{33.6 } & \textbf{29.8 } \\
    \midrule
    \multicolumn{13}{c}{\textit{Open-source MLLMs (Zero-shot)}} \\
    \midrule
    LLaMA-Adapter-V2-7B~\cite{gao2023llama} & 23.9  & 21.2 & \textbf{25.5}  &$ 11.3$  & $32.3$  &$ 31.8$  & $5.7$   & $6.2 $  & 5.9  & $6.1$   & $4.2$   & $6.1$  \\
    TinyLLaVA-3B~\cite{DBLP:journals/corr/abs-2402-14289} & 26.7  & 20.4  & 19.7  & \textbf{23.1}  & 44.9  & 31.8  & \textbf{12.0} & \textbf{13.2 } & \textbf{13.5 } & \textbf{13.3} & \textbf{12.8} & 7.0  \\
    LLaVA-1.5-7B~\cite{DBLP:journals/corr/abs-2310-03744} & 20.0  & \textbf{22.7}  & 7.7   & 11.8  & 26.6  & 33.0  & 8.4   & 7.0   & 7.4   & 8.2   & 8.8   & \textbf{10.8 } \\
    LLaVA-1.5-13B~\cite{DBLP:journals/corr/abs-2310-03744} &   \textbf{26.8}   &    19.7   &   20.2    &   18.8    &   \textbf{45.6}   &  \textbf{36.9}     & $7.6$   & $8.8$   & $7.6 $  & $7.4$   & $7.4$   &$ 6.9$  \\
    \midrule
    \multicolumn{13}{c}{\textit{Models Fine-tuned on LLaVA-1.5-7B}} \\
    \midrule
    Direct (question --> answer) & 26.1  & 24.5  & 26.4  & 14.5  & 36.7  & 30.7  & 9.4   & 10.9  & 11.9  & 12.4  & 8.1   & 3.7  \\
    CoT~\cite{DBLP:conf/acl/HsiehLYNFRKLP23}   & 25.8  & 21.9  & 23.1  & 21.5  & \textbf{37.3 } & 29.1  & 7.3   & 7.6   & 6.5   & 7.1   & 7.1   & 8.4  \\
    CoT-T (filter out incorrect CoT) & 27.5  & 26.0  & 26.9  & 21.0  & 36.1  & 29.6  & 13.7  & 15.7  & 13.7  & 14.0  & 14.5  & 10.5  \\
    CoT-GT~\cite{li2022explanations} & 28.8  & 27.1  & 31.2  & 22.6  & 31.6  & \textbf{32.4 } & 14.0  & 14.6  & 15.5  & 15.5  & 15.0  & 9.3  \\
    T-SciQ (LM)~\cite{DBLP:journals/corr/abs-2305-03453}  & 24.6  & 26.8  & 14.4  & 15.1  & 36.7  & \textbf{32.4 } & 6.8   & 7.6   & 4.9   & 5.3   & 5.6   & 10.5  \\
    T-SciQ (MM)~\cite{DBLP:journals/corr/abs-2305-03453}  & 27.0  & 28.6  & 14.9  & 26.3  & 36.1  & 31.3  & 4.9   & 6.3   & 5.6   & 4.8   & 4.6   & 3.2  \\
    \textbf{VCAR (our method)}  & \textbf{33.7 } & \textbf{30.9 } & \textbf{34.6 } & \textbf{38.7 } & \textbf{37.3 } & 28.5  & \textbf{16.2 } & \textbf{17.1 } & \textbf{16.8 } & \textbf{16.0 } & \textbf{17.0 } & \textbf{14.1 } \\
    \midrule
    \multicolumn{13}{c}{\textit{Models Fine-tuned on TinyLLaVA-3B}} \\
    \midrule
    Direct (question --> answer) & 31.8  & \textbf{30.1 } & 26.4  & 32.3  & 35.4  & 36.9  & 15.1  & 16.5 & 16.5  & 15.4  & 15.1  & 12.2  \\
    CoT~\cite{DBLP:conf/acl/HsiehLYNFRKLP23}   & 30.3  & 20.8  & 29.3  & 32.3  & 38.6  & 36.3  & 6.2   & 5.5   & 4.9   & 4.7   & 5.1   & 10.7  \\
    CoT-T (filter out incorrect CoT) & 30.0  & 26.4  & 23.1  & 30.6  & 40.5  & 33.5  & 14.2  & 15.6  & 14.5  & 15.4  & 15.2  & 10.2  \\
    CoT-GT~\cite{li2022explanations} & 32.7  & 26.0  & 32.2  & 36.0  & 37.3  & 35.8  & 16.6  & \textbf{18.9 } & 16.4  & 17.5  & 17.0  & 13.0  \\
    T-SciQ (LM)~\cite{DBLP:journals/corr/abs-2305-03453} & 29.6  & 29.0  & 30.3  & 19.9  & 36.7  & 33.5  & 13.1  & 15.4  & 12.4  & 11.8  & 12.2  & \textbf{13.8 } \\
    T-SciQ (MM)~\cite{DBLP:journals/corr/abs-2305-03453} & 31.0  & 26.0  & 25.5  & 31.2  & \textbf{39.9}  & 36.9  & 15.1  & 17.5  & 15.5  & 15.4  & 14.1  & 12.8  \\
    \textbf{VCAR (our method)}  & \textbf{34.6 } & 28.6  & \textbf{35.6 } & \textbf{34.4 } & \textbf{39.9}  & \textbf{38.0 } & \textbf{16.7 } & 14.3  & \textbf{17.9 } & \textbf{18.5 } & \textbf{19.7 } & 13.1  \\
    \bottomrule
    \end{tabular}%
    }
    \caption{Experimental results on MathVista and MathVerse. MathVista includes 5 types of math problems: figure question answering (FQA), geometry problem solving (GPS), math word problem (MWP), textbook question answering (TQA), and visual question answering (VQA). MathVerse problems are categorized according to the degree of information content in multi-modality, including text-dominant (TD), text-lite (TL), visual-intensive (VI), visual-dominant (VD), and visual-only (VO). 
    Highest scores within each category are marked as \textbf{Bold}. }
  \label{tab:main}%
\end{table*}%

\subsection{Two-step Training Pipeline}
After acquiring the image descriptions and rationales, we can proceed with the training for both visual comprehension and mathematical reasoning abilities. 
Given the training sample $(v,q,d,r,y)$, one straightforward approach is to train the model to generate both the description and rationale before reaching the final answer, leading to the loss function below:
\begin{equation*}
    \begin{aligned}
        \mathcal{L}&=-\log P(y,r,d|q,v) \\
                    &=-\log P(y|r,d,q,v)P(r|d,q,v)P(d|q,v).
    \end{aligned}
\end{equation*}
Benefiting from the auto-regressive decoding of LLMs, these conditional probabilities above
can be calculated in a single forward pass.
Nevertheless, we found that this joint optimization process leads to sub-optimal performance, which could be attributed to the imbalanced learning for visual comprehension and mathematical reasoning.
Regarding this, we propose to separate the training process by modeling the two capabilities through independent parameters. In experiments of this work, we implement it by exploiting two sets of LoRA~\citep{DBLP:conf/iclr/HuSWALWWC22} modules.




\noindent \textbf{Visual Comprehension Training}. 
The first set of LoRA modules is trained to learn the visual comprehension capability through description generation, which, as mentioned above, is optimizing the following objective:
\vspace{-0.3cm}
\begin{equation*}
    \mathcal{L}_d=-\log P\left(d|v, q;\theta_d\right),
    \vspace{-0.3cm}
\end{equation*}
where $\theta_d$ refers to the parameters of the LoRA module for visual comprehension. 

\noindent \textbf{Mathematical Reasoning Training}. 
We utilize the rationale generation task to improve mathematical reasoning ability. The visual description $d$ is integrated as a supplement, which provides clear text-form context, reducing dependence on visual comprehension in the mathematical reasoning training. 
We incorporate another set of LoRA modules with trainable parameters $\theta_r$  for optimization with the following objective:
\vspace{-0.3cm}
\begin{equation*}
    \mathcal{L}_{r} = - \log P\left(r,y|v, q,d; \theta_r\right).
    \vspace{-0.3cm}
\end{equation*}

\noindent \textbf{Inference}.
During the inference, we first enable $\theta_d$ to generate a targeted and precise visual description $\hat{d}$ for the question in the first step. Then we switch to $\theta_c$ for subsequent mathematical reasoning based on the question, the image, and the generated description.


\section{Experiments}
To verify the effectiveness of the proposed method, we conduct extensive experiments on two benchmarks. In this section, we first introduce the experimental settings, followed by the performance comparison and analysis of our method.

\subsection{Setup}
\label{sec:setup}
\textbf{Data}.
We assess the effectiveness of our pipeline on two benchmarks, namely MathVista~\cite{DBLP:journals/corr/abs-2310-02255} and MathVerse~\cite{zhang2024mathverse}. Both of them are multimodal benchmarks that focus on mathematical reasoning. We use the \textit{minitest} split of the two benchmarks, which are composed of 1,000 and 3,940 testing samples, respectively. Since these benchmarks do not provide a corresponding training set, we construct a training set from five existing datasets, namely DVQA~\cite{DBLP:conf/cvpr/KaflePCK18}, ChartQA~\cite{DBLP:conf/acl/MasryLTJH22}, Geometry3K~\cite{DBLP:conf/acl/LuGJQHLZ20}, TabMWP~\cite{DBLP:conf/iclr/Lu0CWZRCK23}, and IconQA~\cite{DBLP:conf/nips/LuQCXZZYLZ21}. These datasets cover a wide range of domains (\eg chart plot, geometric graph, abstract diagram, and tabular image) in multimodal mathematical reasoning. We randomly select 1,000 samples from the training split of each dataset, resulting in a total of 5,000 training samples. The image description and rationales used in training (\S\ref{sec:supervision_collection}) are collected with Gemini-Pro~\cite{DBLP:journals/corr/abs-2403-05530} unless otherwise specified.

\begin{table*}[!t]
  \centering
  
    \resizebox{\textwidth}{!}{
    \begin{tabular}{l|c|c|ccccc|c|ccccc}
    \toprule
    \multirow{2}[4]{*}{\textbf{Model}} & \multirow{2}[4]{*}{\textbf{Method}} & \multicolumn{6}{c|}{\textbf{MathVista}}       & \multicolumn{6}{c}{\textbf{MathVerse}} \\
\cmidrule{3-14}          &       & \textbf{ALL} & \textbf{FQA} & \textbf{GPS} & \textbf{MWP} & \textbf{TQA} & \textbf{VQA} & \textbf{ALL} & \textbf{TD} & \textbf{TL} & \textbf{VI} & \textbf{VD} & \textbf{VO} \\
    \midrule
    \multicolumn{1}{l|}{\multirow{5}[4]{*}{LLaVA}} & \textbf{VCAR} & \textbf{33.7} & \textbf{30.9} & \textbf{34.6} & \textbf{38.7} & 37.3  & 28.5  & \textbf{16.2} & 17.1  & \textbf{16.8} & 16.0  & \textbf{17.0} & \textbf{14.1} \\
\cmidrule{2-14}          & w/o Rationale & 30.3  & 29.4  & 29.3  & 28.0  & 41.8  & 25.1  & 14.8  & 16.4  & 16.1  & \textbf{17.0} & 15.4  & 9.3 \\
          & w/o Description & 28.8  & 27.1  & 31.2  & 22.6  & 31.6  & \textbf{32.4} & 14.0  & 14.6  & 15.5  & 15.5  & 15.0  & 9.3 \\
          & Concatenation Training & 32.5  & 29.7  & 31.2  & 28.5  & \textbf{44.3} & 31.8  & 15.4  & \textbf{17.4} & 16.5  & 15.2  & 16.9  & 10.8 \\
          & Multi-task Training & 30.8  & 29.4  & 31.2  & 26.9  & 35.4  & \textbf{32.4} & 15.5  & \textbf{17.4} & 16.6  & 16.0  & 16.5  & 10.9 \\
    \midrule
    \multicolumn{1}{l|}{\multirow{5}[4]{*}{TinyLLaVA}} & \textbf{VCAR} & \textbf{34.6} & 28.6  & 35.6  & 34.4  & 39.9  & \textbf{38.0} & \textbf{16.7} & 14.3  & \textbf{17.9} & \textbf{18.5} & \textbf{19.7} & 13.1 \\
\cmidrule{2-14}          & w/o Rationale & 34.1  & \textbf{29.0} & \textbf{36.5} & \textbf{37.6} & 38.0  & 31.8  & \textbf{16.7} & 16.9  & 16.9  & 18.0  & 16.1  & \textbf{15.5} \\
          & w/o Description & 32.7  & 26.0  & 32.2  & 36.0  & 37.3  & 35.8  & 16.6  & 18.9  & 16.4  & 17.5  & 17.0  & 13.0 \\
          & Concatenation Training & 33.9  & 28.3  & 34.6  & 33.9  & 39.9  & 36.3  & 15.9  & 17.1  & 17.8  & 16.9  & 17.0  & 10.7 \\
          & Multi-task Training & 33.2  & 27.5  & 31.7  & 37.1  & \textbf{41.1} & 32.4  & 16.0  & \textbf{19.2} & 16.1  & 17.1  & 15.7  & 12.1 \\
    \bottomrule
    \end{tabular}%
    }
    \caption{Performance of the proposed method compared to different variants described in \S\ref{sec:ablation}.}
  \label{tab:ablation}%
\end{table*}%

\noindent \textbf{Implementation Details}. 
 We experiment with two foundation MLLMs: LLaVA-7B~\cite{DBLP:journals/corr/abs-2310-03744} and TinyLLaVA-3.1B~\cite{DBLP:journals/corr/abs-2402-14289}. 
 Due to constraints of computational resources, we train all models with LoRA~\cite{DBLP:conf/iclr/HuSWALWWC22} whose rank is set to $16$ and $8$ for LLaVA and TinyLLaVA, respectively. The learning rate is set to 2e-4 with a warmup during the first 3\% steps. The batch size is set to $4$ and $8$ for LLaVA-7B and TinyLLaVA-3.1B, respectively. All models, if not specified, are trained for 1 epoch on the training set under the same hyperparameters. 

\noindent \textbf{Evaluation}.
We use accuracy as the evaluation metric. All models are evaluated under a zero-shot generative setting: Given the question and the image, the model is prompted to generate the image description and then the rationale.
We use greedy decoding to ensure deterministic results.
For answer extraction, after the model generates the rationale, we additionally prompt it with ``\textit{So the final answer is}''\footnote{This additional turn is also included in model training.} and use regular expressions to extract the answer from this final response.


\noindent \textbf{Baselines.}
We compare VCAR with other common methods to construct the fine-tuning data.
\begin{itemize}
[noitemsep,topsep=1pt,parsep=1pt,partopsep=0pt, leftmargin=*]

    \item \textbf{Direct}: Training models to directly predict the answer without generating any rationales.
    \item \textbf{CoT} (Chain-of-Thought): Similar to~\citet{DBLP:conf/acl/HsiehLYNFRKLP23}, given the image $v$ and question $q$, we ask Gemini to produce a rationale $r$ for model training. It is worth noting that Gemini-generated rationales may contain inaccuracies.
    
    \item  \textbf{CoT-T}: 
    Building on the CoT approach, this variant filters out instances where the final answer predicted in $r$ does not match the gold answer.
    \item \textbf{CoT-GT}: 
    To improve the accuracy of the constructed rationale, this variant includes the gold answer $y$ when prompting Gemini to produce $r$. This variant employs the identical prompt as described in \S\ref{sec:supervision}.
    \item \textbf{T-SciQ(LM)} and \textbf{T-SciQ(MM)}~\cite{DBLP:journals/corr/abs-2305-03453}: 
    These methods deconstruct the training sequence into three components: \textit{skills} necessary to answer the question, a \textit{plan} outlining the reasoning steps, and a \textit{rationale} to derive the final answer. We test both the language model version\footnote{The original paper only used language models for data collection.} (LM) and the multimodal version (MM) of Gemini-Pro to gather data in the T-SciQ format.
    

\end{itemize}

\subsection{Main Results}
Results on MathVista and MathVerse are presented in Table~\ref{tab:main}. We highlight the following findings:

\noindent \textbf{VCAR significantly outperforms all baseline methods across both benchmarks}. This underscores the effectiveness of VCAR in boosting MLLMs' abilities in multimodal mathematical reasoning. Moreover, the consistent enhancement observed across two base models proves the broad applicability of the VCAR approach. 

\noindent \textbf{VCAR shows marked improvements in categories that demand stronger visual understanding abilities}. 
In MathVerse, examples in ``visual-dominant'' and ``visual-only'' categories require models predominantly leverage visual interpretation skills for solving problems. Here, VCAR shows an average of 14.6\% relative improvements over the best baseline, including a 34.4\% relative increase on LLaVA across the ``visual-only'' examples. This is in contrast to an average 5.3\% improvement observed in text-dominant instances. These results indicate that VCAR enhances the visual comprehension abilities of MLLMs, leading to remarkable performance gains in scenarios requiring intensive visual perception.


\noindent \textbf{Visual comprehension is a greater bottleneck than mathematical reasoning in multimodal math tasks}. While the baseline method T-SciQ aims to improve model training by intricately designing a three-step format for the math reasoning rationale, VCAR outstrips T-SciQ across both benchmarks.
This suggests that deficiencies in visual understanding play a critical role in limiting MLLMs' performance, which are not adequately addressed by solely refining the design of the reasoning rationale.

\begin{figure}[!t]
    \centering
    \includegraphics[width=\linewidth]{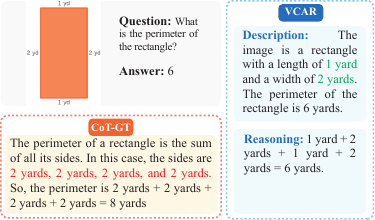}
    \caption{Inference results of VCAR and CoT-GT baseline on an example from MathVista. Correct and incorrect image descriptive expressions are highlighted in green and red, respectively.}
    \label{fig:case}
\end{figure}

\subsection{Ablation Study}
\label{sec:ablation}
To verify the effectiveness of each component in the proposed framework, we compare VCAR with the following variants.
\begin{itemize}
[noitemsep,topsep=1pt,parsep=1pt,partopsep=0pt]

    \item \textbf{w/o Rationale} and \textbf{w/o Description}: In these two variants, we ablate the rationale or the visual description in training and rely solely on the remaining element to derive the answer.
    \item \textbf{Concatenation Training}: Instead of the proposed two-step training, we concatenate the visual description and the reasoning rationale into one response $[d;r]$ to train the model. 
    \item \textbf{Multi-task Training}: We train visual description generation and rationale generation simultaneously, using task-specific prompts. 
\end{itemize}

As indicated in Table~\ref{tab:ablation}, experiments reveal that: 1) The exclusion of any component in VCAR results in reduced performance, underscoring that each component, including the description generation, the rationale generation, and the two-step training pipeline, is indispensable in terms of VCAR's significant improvement.
2) The omission of description notably impairs performance more than the absence of rationales, indicating that inadequate visual understanding is a critical bottleneck in multimodal mathematical reasoning. VCAR effectively addresses this issue through explicit visual description training.

\subsection{Case Study}
To get an intuition of VCAR's effect, we present an example from MathVista in Figure~\ref{fig:case}. The baseline method {CoT-GT} fails to accurately recognize the values depicted in the input image, leading to an incorrect calculation and answer (\textit{2 yards + 2 yards + 2 yards + 2 yards = 8 yards}). In contrast, VCAR enhances the model's visual understanding, enabling it to describe key information in the image precisely. Consequently, the model accurately computes the correct answer (\textit{1 yard + 2 yards + 1 yard + 2 yards = 6 yards}) to the posed question. This instance underscores the importance of accurate visual information in solving multimodal math problems. A similar pattern is observed in the example in Figure~\ref{fig:intro}.

\begin{table}[!t]
  \centering
  \resizebox{\linewidth}{!}{
    \begin{tabular}{l|c|c|ccccc}
    \toprule
    \textbf{Data Source} & \textbf{Method} & \textbf{ALL} & \textbf{FQA} & \textbf{GPS} & \textbf{MWP} & \textbf{TQA} & \textbf{VQA} \\
    \midrule
    \multirow{2}[2]{*}{ Gemini-Pro} & CoT-GT & 28.8  & 27.1  & 31.2  & 22.6  & 31.6  & \textbf{32.4 } \\
          & VCAR  & \textbf{33.7 } & 30.9  & \textbf{34.6 } & \textbf{38.7 } & 37.3  & 28.5  \\
    \midrule
    \multirow{2}[2]{*}{ GPT-4V} & CoT-GT  & 30.5  & 29.4  & 28.4  & 31.2  & 36.7  & 28.5  \\
          & VCAR  & \textbf{33.1 } & 29.7  & \textbf{29.3 } & \textbf{39.8 } & 39.2  & \textbf{30.2 } \\
    \bottomrule
    \end{tabular}%
    }
    \caption{VCAR and CoT-GT with different data sources on MathVista.}
  \label{tab:4v}%
\end{table}%

\section{Analysis}
\label{sec:analy}

\subsection{Using GPT-4V to Collect Data}
\label{sec:4vdata}
To demonstrate the universality of our method, we employ GPT-4V (\textit{gpt-4-vision-preview}) to generate training data in addition to Gemini-Pro. We compare VCAR with the strongest baseline, CoT-GT. 
Table~\ref{tab:4v} presents the comparison results using two data sources. Our method consistently outperforms CoT-GT, irrespective of the data source—GPT-4V or Gemini-Pro. However, the model performance slightly declines compared to training with Gemini-generated data. One possible reason is that GPT-generated responses are longer and more complex, which poses challenges for smaller MLLMs (e.g., LLaVA-7B) to learn. 
Detailed data collection methods with GPT-4V and statistics comparisons of two data sources are provided in Appendix~\ref{apppendix:datasource}.
%

\subsection{The Format of Visual Information}
\label{sec:augques}
\begin{figure}[!t]
    \centering
    \includegraphics[width=\linewidth]{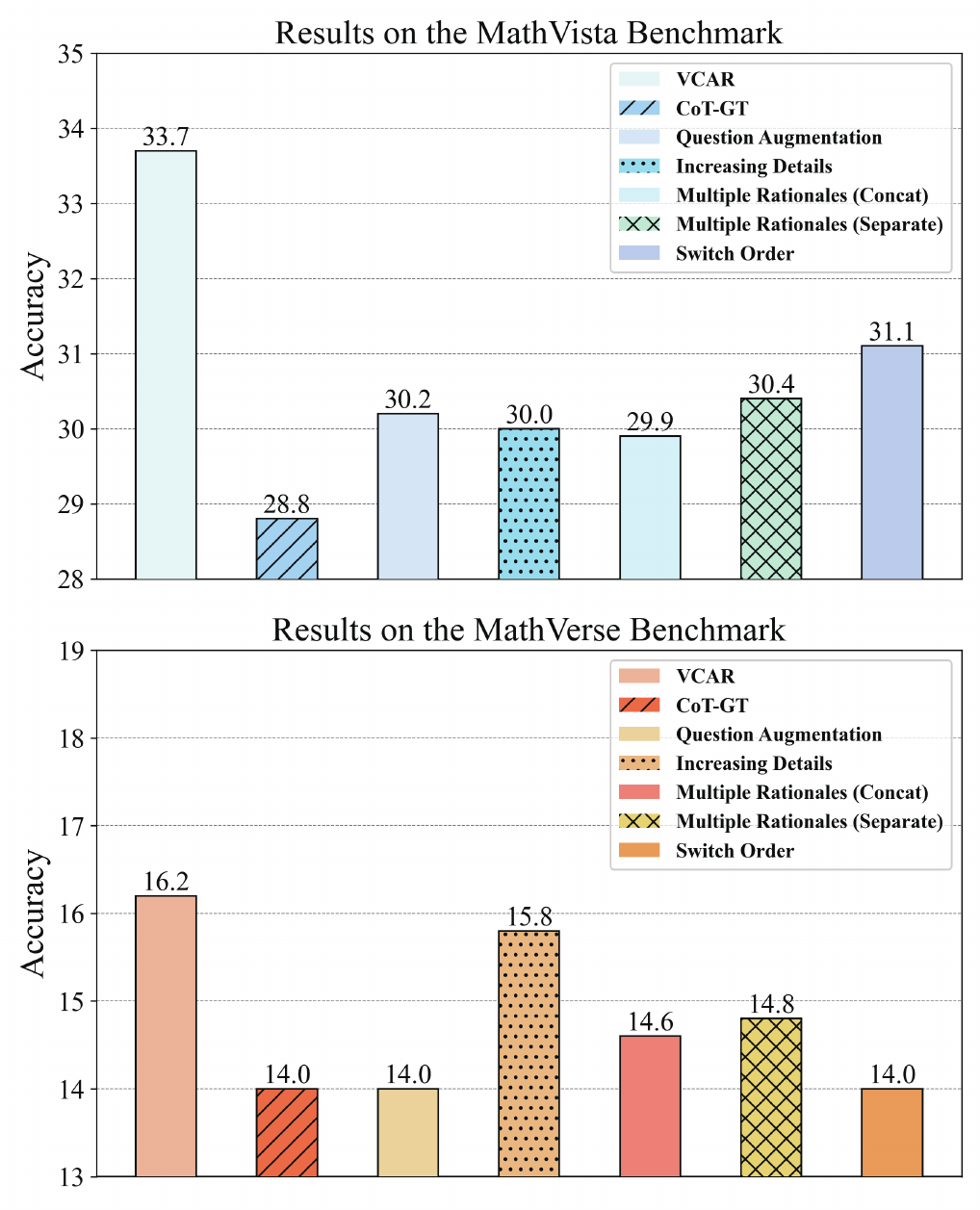}
    \caption{Performance comparisons of various variants on two benchmarks. We present the averaged accuracy. Detailed category-specific results are at Appendix~\ref{apdx:ana}.}
    \label{fig:analyse}
\end{figure}

Constructing image descriptions is an intuitive and direct way of converting visual content into textual modality~\cite{radford2021learning,yarom2024you}.
To explore alternative visual information formats, we substitute the image description in VCAR crafting \textit{visual-augmented questions}. Specifically, we prompt Gemini to rephrase the question by including pertinent visual details observed in the image. The prompt used for question augmentation is demonstrated in the Appendix~\ref{apppendix:augques}.
As depicted in the ``Question Augmentation'' category in Figure~\ref{fig:analyse}, incorporating visual information into questions improves over the CoT-GT baseline. This once again validates that relying solely on rationales leads to insufficient visual understanding supervision.
However, such question augmentation cannot match the improvement brought by VCAR.
This discrepancy is likely due to the image description task being more consistent with the alignment objective during MLLM pre-training. 


\subsection{Impact of Training Sequence Length}
To rule out the influence of increased sequence lengths due to the inclusion of visual descriptions, we develop variants that directly extend the length of reasoning rationales.

\begin{itemize}
\item \textbf{Increasing Details in Rationales}.
\label{sec:ana_extend}
We deliberately prompt Gemini-Pro to incorporate more details based on the previously derived rationale, resulting in a longer version. The specific prompt used is detailed in the Appendix~\ref{apppendix:extend-detail}.
\item \textbf{Multiple Rationales}.
Previous research suggests that aggregating multiple rationales improves performance~\cite{DBLP:conf/acl/ShridharSS23}. Accordingly, we consider using two distinct rationales as another way to expand the number of training tokens. In practice, we prompt Gemini-Pro twice with a sampling temperature of $0.7$.  We explored two training configurations: (1) concatenating the rationales into a single training instance, and (2) pairing each rationale with the input question to create multiple training instances.
\end{itemize}

Figure~\ref{fig:analyse} presents the comparative results of the aforementioned three variants, namely \textit{Increasing Details}, \textit{Multiple Rationales (Concat)}, and \textit{Multiple Rationales (Separate)}. Notably, all three methods fail to yield the same performance as VCAR, which suggests that the improvements brought by VCAR are not merely due to the increased number of training tokens. 

\subsection{Order of the Training Pipeline}
VCAR incorporates a sequential two-step training pipeline. Initially, the pipeline generates image descriptions, followed by the generation of rationales that derive the answer. To assess the necessity of this order, we introduce a variant where the order of the pipeline is reversed: the model first constructs rationales, subsequently generating image descriptions before deriving the final answer.

The ``\textit{Switch Order}'' category results in Figure~\ref{fig:analyse} illustrate that modifying the sequence order adversely impacts performance. This outcome is intuitive as the accurate visual details provided by the image descriptions can assist the rationale generation process. Altering the order obstructs the use of these descriptions in generating rationales.


\section{Conclusion}
In this work, we introduce VCAR, a two-step training pipeline designed to enhance multimodal mathematical reasoning via targeted visual comprehension training. VCAR enhances MLLMs with visual comprehension through image description generation training in the first step. Thereafter, descriptions are incorporated in the second rational generating step to assist in executing mathematical reasoning. We conduct comprehensive experiments across two widely used benchmarks, demonstrating that VCAR outperforms approaches without specified visual comprehension training. Notably, VCAR achieves significant performance improvements on problems requiring a high demand of visual understanding, underscoring the benefits of specialized visual comprehension training in multimodal mathematical reasoning.


\section*{Limitations}

To our knowledge, this work has the following limitations:
\begin{itemize}
[noitemsep,topsep=1pt,parsep=1pt,partopsep=0pt]
\item Due to our limited computational resources, we train all models, including the baselines, with LoRA to maintain a fair comparison. As a result, our two-step training pipeline is implemented with training separate LoRA modules. In future work, we plan to extend our approach to include experiments on full model fine-tuning.
\item As mentioned in \S\ref{sec:setup}, our training data originate from the \textit{training} splits of 5 published datasets. Meanwhile, a part of the \textit{test} splits of these 5 datasets are included in the MathVista benchmark. We adopt the training splits of these datasets due to the scarcity of available training data in the field of multimodal math reasoning, while utilizing stronger models like GPT-4V to generate new large-scale data is beyond our budget. However, this setting remains equitable since all models are trained on the same training instances and there is no instance overlap between the training and test splits of these datasets. We leave the exploration of augmenting high-quality multimodal math data as future work.
\end{itemize}

%

\nocite{PLUG}
\nocite{Auto-instruct}

\section*{Acknowledgements}
This work was supported by NSF IIS-2119531, IIS-2137396, IIS-2142827, IIS-2234058, CCF-1901059, and ONR N00014-22-1-2507.

\bibliography{custom}

\appendix

\newpage
\begin{table}[!t]
  \centering
  \resizebox{\linewidth}{!}{
    \begin{tabular}{l|c|c}
    \toprule
    \textbf{Source} & \textbf{Type} & \textbf{Averatge \#Token} \\
    \midrule
    \multirow{2}[2]{*}{Gemini-Pro} & Description & 48.92 \\
          & Rationale & 55.39 \\
    \midrule
    \multirow{2}[2]{*}{GPT-4V(ision)} & Description & 138.85 \\
          & Rationale & 101.57 \\
    \midrule
    \multirow{2}[2]{*}{GPT-4V(ision) (Disentangle)} & Description & 60.20 \\
          & Rationale & 81.43 \\
    \bottomrule
    \end{tabular}%
    }
  \caption{Statistics of data from two sources. GPT-4V tends to produce relatively lengthy responses compared to Gemini-Pro, which places challenges for training. Using the disentangle prompt alleviates this phenomenon.}
  \label{tab:statistic}
\end{table}%
\section{Detailed Prompts}
\subsection{Question Augmentation Prompt}
\label{apppendix:augques}
We provide the prompt used to augment the question (detailed in Section~\ref{sec:augques}) as follows.
\begin{figure}[h]
    \centering
    \includegraphics[width=\linewidth]{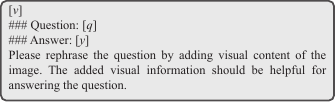}
\end{figure}

\subsection{Increasing Detail Prompt}
\label{apppendix:extend-detail}
Based on a previously obtained rationale $s$, we use the following prompt to deliberately incorporate more details  (detailed in Section~\ref{sec:ana_extend}) as follows.
\begin{figure}[h]
    \centering
    \includegraphics[width=\linewidth]{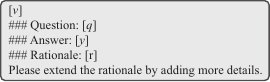}
\end{figure}

\begin{table*}[!t]
  \centering
  \resizebox{\textwidth}{!}{
    \begin{tabular}{l|c|ccccc|c|ccccc}
    \toprule
    \multirow{2}[4]{*}{\textbf{Method}} & \multicolumn{6}{c|}{\textbf{MathVista}}       & \multicolumn{6}{c}{\textbf{MathVerse}} \\
\cmidrule{2-13}          & \textbf{ALL} & \textbf{FQA} & \textbf{GPS} & \textbf{MWP} & \textbf{TQA} & \textbf{VQA} & \textbf{ALL} & \textbf{TD} & \textbf{TL} & \textbf{VI} & \textbf{VD} & \textbf{VO} \\
    \midrule
    VCAR  & \textbf{33.7 } & \textbf{30.9 } & \textbf{34.6 } & \textbf{38.7 } & 37.3  & 28.5  & \textbf{16.2 } & 17.1  & \textbf{16.8 } & 16.0  & 17.0  & \textbf{14.1 } \\
    \midrule
    \textbf{Variants} &       &       &       &       &       &       &       &       &       &       &       &  \\
    \midrule
    Question Augmentation & 30.2  & 29.7  & 31.7  & 24.2  & 37.3  & 29.1  & 14.0  & 14.3  & 14.9  & 14.1  & 16.4  & 10.2  \\
    Increase Details in Ratinales & 30.0  & 29.7  & 25.5  & 22.6  & 39.9  & \textbf{34.6 } & 15.8  & 15.9  & 16.0  & \textbf{16.9 } & \textbf{17.8 } & 12.6  \\
    Rationales Ensemble (Concat) & 29.9  & 27.5  & 25.5  & 29.0  & 40.5  & 30.2  & 14.6  & \textbf{17.3 } & 14.6  & 14.8  & 14.7  & 11.5  \\
    Rationales Ensemble (Separate) & 30.4  & 29.7  & 28.4  & 31.7  & 36.1  & 27.4  & 14.8  & 15.7  & 15.2  & 14.0  & 16.9  & 12.1  \\
    Switch order & 31.1  & 26.8  & 30.8  & 22.6  & \textbf{45.6 } & 34.1  & 14.0  & 14.5  & 14.3  & 14.5  & 15.6  & 11.0  \\
    \bottomrule
    \end{tabular}%
    }
      \caption{Comparison among several variant methods. We highlight the best performances in \textbf{Bold}.}
  \label{tab:detailana}%
\end{table*}%

\section{Statistics Comparison across Data Sources}
\label{apppendix:datasource}
We detail the supervision collection process with GPT-4V in Section~\ref{sec:4vdata}.
Notably, GPT-4V tends to produce lengthy responses that intermingle image descriptions with rationales. To collect targeted responses that solely focus on description or step-wise reasoning that is suitable for VCAR, we tailor the prompt by asking GPT-4V to disentangle these two parts into separate paragraphs. We illustrate the prompt as follows.
\begin{figure}[h]
    \centering
    \includegraphics[width=\linewidth]{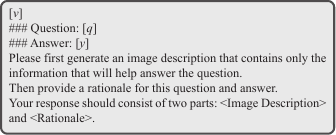}
    \vspace{-1cm}
\end{figure}
\newpage
The statistics comparison between two data sources, namely Gemini-Pro and GPT-4V, is provided in Table~\ref{tab:statistic}

\section{Detailed Results of Variants}
\label{apdx:ana}
We present the category-specific results of the variants in Section~\ref{sec:analy} in Table~\ref{tab:detailana}.

\end{document}